\documentclass[10pt,twoside]{article}
\usepackage[latin1]{inputenc}
\usepackage[T1]{fontenc}
\usepackage[english,francais]{babel}
\usepackage{graphicx}
\usepackage{hyperref}
\hypersetup{colorlinks=true, linkcolor=blue, filecolor=blue, pagecolor=blue, urlcolor=blue}

\def\auteur{Bernard \textsc{Jacquemin}\up{1} et Sabine \textsc{Ploux}\up{2}}
\def\hbpauteur{Bernard \textsc{Jacquemin} et Sabine \textsc{Ploux} (2008)}
\def\adresselabo{\up{1}LIMSI CNRS UPR 3251 -- Orsay (France)\\
                 \up{2} UCB Lyon 1 et L2C2 CNRS UMR 5230 -- Bron (France)}

\def\courriel{\href{Bernard.Jacquemin@limsi.fr}{Bernard.Jacquemin\,@\,limsi.fr}\\\href{sploux@isc.cnrs.fr}{sploux\,@\,isc.cnrs.fr}}

\def\titre{Du corpus au dictionnaire.\\Réalisation automatique d'un outil de gestion de
l'information multilingue}

\def\mabiblio{bj}

\def\titrecourt{Du corpus au dictionnaire}

\def\piedpage{\textit{Cahiers de Linguistique}, 33(1), pp. 63--84.}

\usepackage{fancyhdr}
\title{\titre}
\author{\auteur\\\adresselabo\\\courriel}
\date{}

\begin{document}

\maketitle

\begin{abstract}
\noindent Dans cet article, nous proposons une méthode automatique de construction de ressources lexico-sémantiques multilingues pour naviguer par le sens à travers l'information contenue dans des bases textuelles de langues différentes. Cette méthode s'appuie sur un modèle mathématique de représentation du sens appelé Atlas sémantiques, qui consiste à exploiter des relations linguistiques entre des unités lexicales pour construire des graphes, projetés dans un espace sémantique qui constitue une carte dénotant les tendances de sens d'un mot considéré. À partir de l'analyse morpho-syntaxique d'un corpus, et en utilisant les relations syntaxiques entre les items du corpus, il est possible de constituer une ressource lexico-sémantique qui décrit l'ensemble des sens attestés dans le corpus pour tout le lexique qui y est représenté, grâce aux contextes syntaxiques typiques des entrées décrites. Il est également possible de conserver un lien systématique entre les tendances de sens représentées et les énoncés qui ont servi à les construire, et donc de relier toutes les instances d'un mot dans un sens donné pour naviguer entre elles. Il est également possible, en utilisant des corpus de langues différentes, de construire des ressources qui se correspondent entre langues, et de naviguer entre les textes grâce à la traduction, même partielle, des contextes syntaxiques.\\
\textbf{Mots-clefs:} ressource lexico-sémantique, représentation du sens, navigation sémantique, gestion d'information multilingue, corpus multilingue, navigation interlangue.
\end{abstract}

\selectlanguage{english}
\begin{abstract}
\noindent In this article, we propose an automatic process to build multi-lingual lexico-semantic resources. The goal of these resources is to browse semantically textual information contained in texts of different languages. This method uses a mathematical model called Atlas sémantiques in order to represent the different senses of each word. It uses the linguistic relations between words to create graphs that are projected into a semantic space. These projections constitute semantic maps that denote the sense trends of each given word. This model is fed with syntactic relations between words extracted from a corpus. Therefore, the lexico-semantic resource produced describes all the words and all their meanings observed in the corpus. The sense trends are expressed by syntactic contexts, typical for a given meaning. The link between each sense trend and the utterances used to build the sense trend are also stored in an index. Thus all the instances of a word in a particular sense are linked and can be browsed easily. And by using several corpora of different languages, several resources are built that correspond with each other through languages. It makes it possible to browse information through languages thanks to syntactic contexts translations (even if some of them are partial).
\end{abstract}
\selectlanguage{francais}

\section{Introduction}
\label{sec:Introduction}

 Dans notre société, la quantité d'information
textuelle disponible augmente de manière telle
qu'aucun être humain ne peut plus la maîtriser. Or
elle est devenue une richesse capitale dans des secteurs aussi variés
que la politique, la culture, l'enseignement, la
défense, l'économie, etc. La situation est
d'autant plus difficile à gérer que
l'information n'est plus
nécessairement disponible dans la langue des utilisateurs, mais
qu'elle est également à chercher en
d'autres langues. Des approches automatiques sont
dès lors nécessaires pour identifier et indexer les contenus afin
d'y donner un accès aisé et immédiat à la
demande.

 Les approches automatiques de la gestion de
l'information se heurtent toutefois à une
difficulté majeure : celle de la compréhension des textes, et des
mots qui les composent. L'utilisation de
dictionnaires, naturellement mise en {\oe}uvre immédiatement, a
dévoilé plusieurs défauts majeurs de ces ouvrages
lorsqu'ils sont utilisés dans le cadre
d'une exploitation automatique : couverture
insuffisante du lexique, découpage en acceptions arbitraire et
parfois incohérent, information souvent lacunaire ou peu
systématique, structure et données aisément compréhensibles
pour un être humain, mais nécessitant des connaissances
préalables et complexes pour la machine... Plusieurs initiatives ont
cherché à pallier ces défauts. Les unes ont amené à la
création de ressources conceptuelles ou ontologiques censées
représenter l'univers selon une hiérarchie (par
exemple \textit{Cyc}, \cite{LenatGuha90}), d'autres
recensent le lexique qu'elles organisent en ensembles
sémantiques (comme \textit{WordNet}, \cite{FellbaumA98}) ou
combinatoires (comme le \textit{DECFC},
cf.~\cite{MelcukAl84,MelcukAl88,MelcukAl92,MelcukAl99}), certaines
tentent de combiner les données disparates fournies par plusieurs
dictionnaires \cite{Jacquemin05}. Cependant, aucune de ces
approches n'a jusqu'à présent
résolu à la fois les problèmes de couverture, de découpage
sémantique et d'objectivité. Par ailleurs, les
difficultés relatives au passage de textes d'une
langue à une autre, bien connues du monde de la traduction,
qu'elle soit humaine, assistée par ordinateur ou
automatique, réclament des ressources toujours plus riches et plus
précises, ce qui rend le problème d'autant plus
complexe.

 Un modèle vise cependant à représenter le sens à partir de
relations d'ordre linguistique entre unités
lexicales grâce à une approche mathématique et statistique
\cite{Ploux97}. Ce modèle, appelé \textit{Atlas sémantiques},
a montré au cours de plusieurs expériences ses capacités à
appréhender la sémantique lexicale. Nous proposons
d'exploiter les qualités de ce modèle en utilisant
comme lien entre les unités lexicales des relations syntaxiques
issues de l'analyse de corpus. La dimension
multilingue est assurée par le choix de corpus distincts, de langues
différentes, de grande taille et de contenu comparables, telles les
différentes instances de l'encyclopédie
\textit{Wikipédia}. Pour chaque langue traitée, nous pensons être
en mesure de constituer une ressource lexico{}-sémantique de
qualité, qui devrait lever une grande partie des réticences
affichées par le domaine du traitement de
l'information à propos de
l'arbitraire dans le découpage choisi et à propos
la couverture, tant lexicale que sémantique, à condition que les
corpus soient suffisamment représentatifs de la langue à traiter.

 Par ailleurs, la construction du dictionnaire à partir
d'un corpus permet de conserver un lien direct entre
l'information sémantique de la ressource et les
énoncés correspondants dans le corpus, qui constituent dès lors
des exemples \textit{in situ} des usages réels. De ce fait, le
dictionnaire construit constitue un excellent moyen de naviguer à
travers l'information contenue dans les corpus en
utilisant le sens comme référence de navigation. Enfin,
l'exploitation d'un dictionnaire de
traduction entre les ressources de langues distinctes permet de mettre
en rapport des espaces sémantiques comparables, propres à chaque
corpus, et de rapprocher non seulement des mots traduits, mais
également des sens et des bribes de textes. Cela constitue de ce fait
un outil d'aide à la traduction apparenté aux
systèmes à mémoire de traduction, mais également une
possibilité de passer d'une information contenue
dans un des corpus à son équivalent dans un autre corpus, et donc
dans une autre langue.

 Nous commençons par présenter le modèle des Atlas
sémantiques, ses qualités, ses spécificités et ses applications
actuelles avant d'exposer la méthode que nous
proposons pour construire une ressource sémantique propre à combler
certaines carences des dictionnaires traditionnels et à fournir un
outil de navigation à travers une information textuelle. Ensuite,
nous montrons comment un dictionnaire de traduction permet de relier
aisément et efficacement nos ressources lexico{}-sémantiques
distinctes. Enfin, nous concluons en présentant les perspectives
ouvertes par ce projet.

\subsection{Le modèle des Atlas sémantiques}
\label{sec:Modele}

\subsection{Représentation du sens et synonymie}
\label{sec:synonymie}

 L'équipe de S. Ploux, \textit{Modèles
mathématiques et neuropsychologiques pour le langage} (L2C2, CNRS), a
mis au point les Atlas sémantiques, un modèle tout à fait
original de représentation du sens très éloigné du découpage
en acceptions propre aux dictionnaires traditionnels. La validité de
la représentation du sens par ce modèle a été montrée dans
des évaluations lexicologiques \cite{Ploux97,PlouxVictorri98} 
et psycholinguistiques (cf.~\cite{RouibahAl01,Ji04} et, 
dans une moindre mesure, \cite{Maslov04}). Il présente entre
autres la particularité d'analyser statistiquement
des liens établis entre unités lexicales par des humains afin
d'établir le plus objectivement possible une carte
représentant les différentes tendances de sens pour chacun des mots
traités. Les différentes tendances sont exprimées visuellement et
relativement, par la projection, dans un espace cartographique, de mots
auquel le mot considéré est relié. Chaque carte permet donc
d'avoir une vision intuitive de la richesse
sémantique d'un mot considéré.

 Le modèle a été originellement conçu pour résoudre les
problèmes relatifs à l'utilisation simultanée
d'indications de synonymie\footnote{Cette application synonymique des Atlas sémantiques est
disponible en ligne sur
\href{http://dico.isc.cnrs.fr/}{http://dico.isc.cnrs.fr.}} fournies par sept
dictionnaires \cite{Guizot1848,Lafaye1861,Benac56,Bailly67,DuChazaud71,Larousse71-77,Robert85}.
En effet, le découpage en acceptions disparate effectué par ces
dictionnaires les rendait impropres à une mise en commun simple des
correspondances synonymiques. De ce fait, \cite{Ploux97} a proposé
d'exploiter la théorie des graphes et
l'analyse factorielle des correspondances pour
concevoir une méthode qui articule les mots les uns par rapport aux
autres sous l'angle du sens. Ce modèle se veut
objectif dans la mesure où il est fondé sur une analyse statistique
des liens de synonymie entre les mots étudiés pour aboutir à une
représentation du sens. Il se veut également intuitif puisque la
distribution des sens est présentée dans un espace géométrique
multidimensionnel, où chaque mot se voit représenté dans une
carte qui lui est propre. Enfin, il manifeste sa différence en
présentant les sens d'un même mot non plus selon
un découpage strict des différentes acceptions, mais dans un
continuum sémantique où la distance entre deux acceptions est
fonction de la différence entre les sens qui leur sont associés.

 Le modèle se fonde sur la construction de graphes à partir de
relations entre entités. Les entrées et leurs synonymes constituent
les entités, les sommets, tandis que le lien de synonymie entre ces
unités lexicales en représente les arêtes. Un seul type de graphe
est conservé dans le cadre du modèle : la \textit{clique}. Il
s'agit d'un type de graphe
particulier dans lequel tous les sommets sont interconnectés les unes
avec les autres, ce qui signifie que chaque unité lexicale
considérée possède une relation de synonymie explicite avec
toutes les autres qui constituent la clique. Ce graphe
particulièrement dense relie donc des unités qui sont très
étroitement liées d'un point de vue sémantique.
L'interconnexion de chaque unité lexicale composant
une clique avec plusieurs autres unités permet de déterminer le
sens particulier de chaque unité dans la clique considérée. On
pourra par exemple retrouver le mot \textit{type} dans deux cliques
très différentes, au voisinage de \textit{amant} ou
\textit{bonhomme} pour l'une, dans un sens lié au
couple, ou avec \textit{exemple} ou \textit{archétype} pour
l'autre, dans une acception dénotant la catégorie.
La clique constitue dès lors un niveau de granularité de sens plus
fin que le mot lui{}-même, plus fin que l'acception
du dictionnaire classique également, car les cliques sont
généralement plus nombreuses que les acceptions, et deux cliques
très voisines peuvent ne varier que par une ou deux unités
lexicales, et recouvrir des significations qui se confondent.

\begin{figure}
 \centering
 \includegraphics[width=110mm]{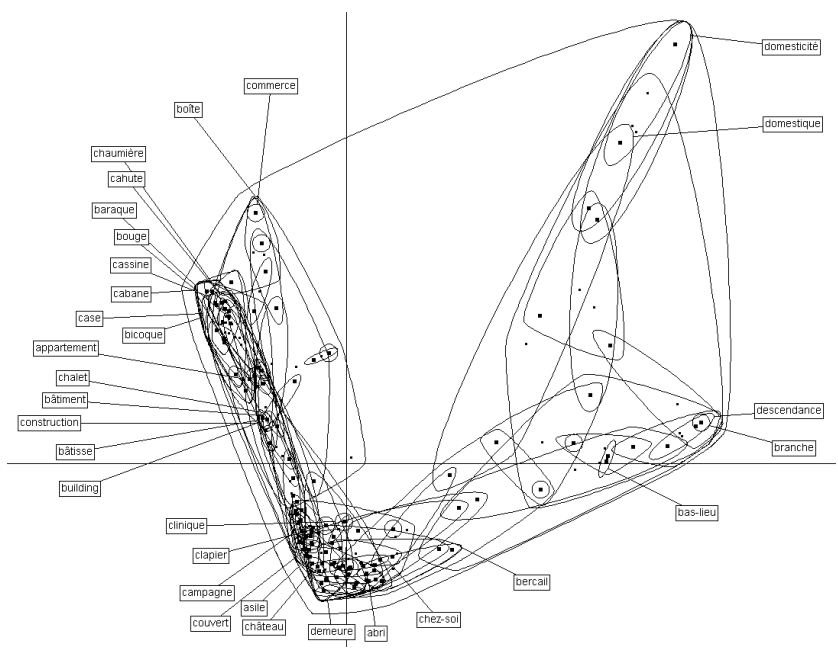}
 \caption{Carte sémantique synonymique de \textit{maison}.}
 \label{fig:maison}
\end{figure}

 Une fois les cliques constituées, un traitement statistique appelé
analyse factorielle des correspondances est appliqué à chacune des
cliques constituées pour une unité donnée. Ce traitement permet
de disposer dans un espace géométrique multidimensionnel chacune
des cliques dont les coordonnées varieront en fonction de son contenu
et de la densité des liens que les différents dictionnaires
établissent entre ces unités lexicales. Une projection de cet
espace multidimensionnel sur un plan en deux dimensions permet de
visualiser les tendances sémantiques du mot considéré,
relativement aux synonymes contenus dans les cliques ainsi
visualisées. La figure~\ref{fig:maison} montre la carte sémantique du mot
\textit{maison}, dont les tendances de sens sont manifestées par des
synonymes, qui distinguent la raison sociale, le bâtiment, la
domesticité, la famille, etc.

 Ce modèle fournit dès lors une ressource sémantique qui
représente objectivement des relations d'ordre
sémantique dont l'établissement
n'est pourtant pas forcément aussi objectif,
puisqu'il est réalisé par les auteurs des
différents dictionnaires utilisés. Cette ressource, qui comporte
l'information extraite de sept dictionnaires,
présente de ce fait une richesse inusitée tout en manifestant une
grande cohérence dans son information. Enfin, le \textit{continuum}
sémantique dans lequel chaque tendance sémantique
s'inscrit brise la structure classique des
dictionnaires traditionnels dont le découpage en acceptions est
souvent arbitraire et trop exclusif.

\subsection{Atlas sémantique et corpus}
\label{sec:corpus}

 La capacité remarquable du modèle des Atlas sémantiques à
décrire le sens lexical à partir de relations de synonymie entre
unités lexicales a amené l'équipe de S. Ploux,
et plus particulièrement H. Ji à s'intéresser
à d'autres applications à travers
d'autres types de liens. En effet,
l'application originelle souffre de son besoin de
disposer de plusieurs dictionnaires préexistants, coûteux en temps
et en argent, pour construire les graphes. De plus,
l'association entre une tendance de sens et des
synonymes n'est pas toujours pertinente dans le cadre
de l'étude cognitive du langage.
C'est donc logiquement que \cite{PlouxJi03} en
sont venus à considérer un corpus comme une ressource pertinente
pour dénoter le sens lexical sans dictionnaire, utilisant pour ce
faire le vocabulaire typiquement associé à un mot pour chacune de
ses tendances de sens plutôt que ses synonymes.

 De fait, un corpus contient intrinsèquement une structure
qu'il est aisé de représenter sous forme de
graphes, à partir desquels peuvent être sélectionnées les
cliques lorsque ces graphes contiennent des sommets qui sont tous
interconnectés. Ainsi, si la relation choisie entre les mots est
l'appartenance à un même contexte, et que le
contexte est défini par une fenêtre prédéterminée, tous les
mots appartenant à cette fenêtre sont interconnectées et
appartiennent virtuellement à la même clique. La nouvelle ressource
issue de cette approche est donc contextuelle\footnote{\cite{JiPloux03} donnent le nom
d'\textit{ACOM} (\textit{Automatic Contexonym
Organizing Model}) à cette application. Pour la clarté de notre
propos, nous parlerons désormais d'\textit{Atlas
sémantique contextuel} parallèlement à
l'\textit{Atlas sémantique synonymique} originel.
Pour les représentations sémantiques visuelles, nous utiliserons
également les termes de \textit{cartes sémantiques contextuelles}
et de \textit{cartes sémantiques synonymiques}, respectivement.} là où le
dictionnaire originel est synonymique. Les contextes typiques qui
apparaissent dans les cartes sémantiques sont appelés
\textit{contexonymes} \cite{Ji04}.

 L'utilisation conjointe du modèle des Atlas
sémantiques et d'un corpus impose toutefois deux
remarques importantes. La première concerne
l'étendue du corpus. En effet, la richesse de la
ressource obtenue est évidemment fonction de la richesse du corpus,
c'est{}-à{}-dire que seuls les mots représentés
dans le corpus sont présents dans la ressource, puisque ce sont ces
mots qui servent à construire la ressource. Pour une raison tout
aussi évidente, seuls les sens attestés dans le corpus peuvent
apparaître dans la ressource qui en est issue, puisque
l'unité de sens est la clique et que les cliques
sont issues de l'analyse du corpus. Le corpus doit
donc être suffisamment conséquent non seulement pour contenir un
lexique jugé de taille raisonnable {--} un dictionnaire général
classique contient pour le français environ 60~000 entrées {--},
mais également pour que les différents sens de chacune des unités
lexicales de ce vocabulaire y soient attestés. Un examen attentif du
corpus, tant dans sa variété lexicale que dans la diversité des
sujets abordés, voire des genres littéraires représentés, est
dès lors nécessaire \cite{BiberAl98}. La seconde remarque
découle logiquement de la première. En effet, la taille
nécessairement importante du corpus ne permet pas
d'envisager l'utilisation simple de
toutes les cliques concernées, dont le nombre et la diversité
seraient sources de bruit. \cite{JiAl03} ont
donc mis en place plusieurs critères de limitation et de contrainte
qui permettent d'augmenter la précision du
résultat. Notamment, il est possible, pour chaque clique construite
pour un mot donné, de limiter les mots considérés comme
pertinents aux seuls dont la fréquence d'apparition
dans le contexte du mot considéré dépasse un seuil prédéfini.
Les contextes rares sont ainsi éliminés. Comme cette contrainte ne
suffit pas à limiter suffisamment le bruit, les contextes de ces mots
proposés pour la construction des cliques sont eux{}-mêmes
étudiés, de manière à supprimer également les contextes où
un cooccurrent fréquent du mot considéré se trouve lui{}-même
dans un contexte qui lui est rare. Dans un même ordre
d'idée, les mots les plus fréquents du corpus sont
également éliminés de la construction des cliques de manière
à éviter la présence systématique d'articles,
prépositions, auxiliaires et autres mots{}-outils qui sont porteur
d'une sémantique faible et peuvent rarement amener
à discriminer les différents sens d'une même
unité lexicale.

\begin{figure}
 \centering
 \includegraphics[width=110mm]{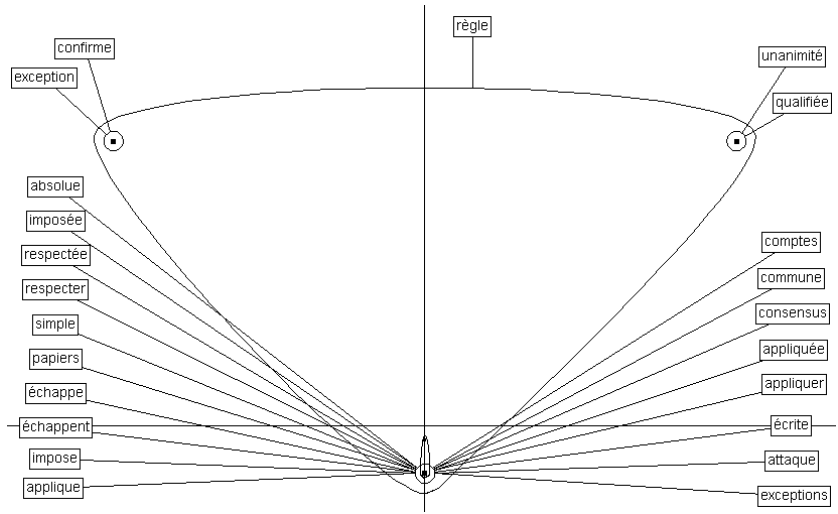}
 \caption{Carte sémantique contextuelle de \textit{règle}.}
 \label{fig:regle}
\end{figure}

 La fenêtre utilisée est soit une fenêtre arbitraire de
vingt{}-cinq mots \cite{Ji04}, soit la phrase \cite{JiPloux03}. 
Les cinq cents mots les plus fréquents du corpus sont
éliminés du calcul des cliques. Ce sont généralement les
unités qui font partie des cinq pour{}-cent des contextes les plus
fréquents qui sont conservées pour construire ces cliques. Une fois
ces dernières construites, une analyse factorielle des
correspondances semblable à celle utilisée dans le cadre de la
synonymie permet de les disposer dans un espace géométrique
multidimensionnel, dont la projection dans un plan représente
objectivement les diverses tendances d'un mot donné.
Cependant, à la différence des espaces synonymiques, ce sont les
contextes les plus typiques de l'unité
considérée dans un sens donné qui permettent
d'en distinguer les différents sens\footnote{Un prototype de cette application du modèle des Atlas
sémantiques à un corpus est consultable en ligne sur
\href{http://dico.isc.cnrs.fr/fr/dico/context/search}{http://dico.isc.cnrs.fr/fr/dico/context/search}.
Les corpus utilisés sont le \textit{British National Corpus} pour
l'anglais et le corpus du journal \textit{Le Monde}
(1997{}-2002) pour le français.}. La
figure~\ref{fig:regle} montre la carte sémantique du mot \textit{règle} issue de
l'examen du corpus. Ce sont dès lors les contextes
typiques du mot qui en indiquent les différentes tendances de sens.

 Dans la perspective d'une étude cognitive du
langage, diverses expériences ont été menées pour tester la
validité de cette application du modèle. Dans
l'une d'entre elles notamment, \cite{Ji04}
compare des associations de mots réalisées sur
présentation d'un mot{}-stimulus par des sujets
humains et les contextes typiques fournis à ces mêmes
mots{}-stimuli par le modèle, et y retrouvent globalement le même
lexique. \cite{Maslov04} a également montré que la construction de
deux ressources distinctes à partir de deux corpus anglais de
domaines différents (\textit{The MIT Encyclopedia of the Cognitive
Sciences} pour l'un et environ 120 résumés
(\textit{abstracts}) d'articles neuroscientifiques
pour l'autre) produit des cartes sémantiques
différentes, représentatives des habitudes langagières des
rédacteurs de l'un ou de l'autre
domaine. Si ces expériences tendent à montrer que le modèle est
capable de créer une cartographie de la perception du sens chez les
rédacteurs d'un corpus ou d'un
dictionnaire, elles montrent également son aptitude à décrire
sans a priori l'ensemble des sens attestés pour tout
le lexique du corpus utilisé. Il est d'ailleurs
remarquable que la carte sémantique contextuelle du mot
\textit{règle} (figure \ref{fig:regle}), obtenue suite à
l'examen du corpus \textit{Le Monde 1997{}-2002}, ne
comporte pas toutes les acceptions de ce terme, et omet notamment les
sens liés à l'instrument de mesure ou aux
menstruations féminines. En effet, ces sens
n'apparaissent pas dans le corpus
d'origine. Dès lors, il est évident que
lorsqu'un corpus spécialisé est utilisé pour
établir un dictionnaire sémantique selon la méthode décrite
plus haut, la ressource qui en découle est elle{}-même un
dictionnaire de spécialité, dont le lexique et la sémantique sera
le fidèle reflet du domaine traité par le corpus, avec les
restrictions déjà présentées dans le choix du corpus
\cite{JacqueminPloux06}.

\section{Sémantique lexicale et navigation informationnelle}
\label{sec:navigation}

\subsection{Contexte syntaxique en corpus et représentation du sens}
\label{sec:contexte}

 Mais si le modèle des Atlas sémantiques est bien apte à recenser
et à représenter les mots et les sens des mots attestés dans un
corpus, la méthode mise en {\oe}uvre pour ce faire comporte des
lacunes ou des imperfections qu'il est nécessaire de
faire disparaître pour augmenter la confiance que
l'on peut avoir dans la ressource produite et pour
diminuer encore le bruit, qui reste important malgré les techniques
statistiques de réduction mentionnées plus haut. Sur la figure~\ref{fig:regle},
la distinction entre \textit{échappe} et \textit{échappent}, de
pure forme, n'est pas nécessaire pour spécifier
des contextes typiques. Les défauts concernent
d'abord la faible qualité de la relation entre les
mots, ensuite la distinction erronée entre les différentes formes
d'une même unité lexicale, enfin la confusion
entre des unités distinctes mais homographes. Ces faiblesses de la
technique employée sont sources d'erreurs et nous
proposons d'y remédier par une approche mixte qui
fait intervenir, en plus des méthodes mathématiques et statistiques
propres au modèle des Atlas sémantiques, plusieurs traitements
d'ordre linguistique.

 Notre première proposition concerne
l'établissement d'un lien entre les
mots plus pertinent que la simple proximité dans une fenêtre
arbitraire, et même plus fort que l'appartenance à
une même phrase. En effet, l'utilisation statistique
d'un seuil de fréquence pour déterminer
l'importance d'un contexte dans la
méthode d'analyse de corpus présentée
ci{}-dessus, quoique très efficace et simple à mettre en {\oe}uvre,
n'en reste pas moins aléatoire et empirique. Nous
proposons donc d'exploiter un type de relation qui
garantit une interdépendance réelle entre les unités lexicales
utilisées comme contextes typiques. Les relations syntaxiques
permettent de garantir une appartenance réelle des différentes
unités au même contexte. En effet, les approches
distributionnalistes {--} par exemple les travaux de \cite{Firth57}
ou de \cite{Harris68} {--} ont montré une proximité de sens
remarquable entre les unités lexicales dont le contexte syntaxique
est similaire. Par ailleurs, la nécessité
d'identifier l'acception correcte en
contexte d'un terme polysémique, dans des
problématiques telles que la traduction automatique \cite{Weaver49} ou
la désambiguïsation sémantique lexicale \cite{Reifler55}, a mis en
évidence l'intérêt du contexte syntaxique dans
ce processus. Dès lors, un système d'analyse
syntaxique va s'inscrire dans la phase
d'examen du corpus, de manière à établir des
liens syntaxiques entre unités lexicales et entre têtes de groupes
syntaxiques. Toutefois, une application trop stricte de cette forme de
contrainte risquerait de faire perdre le bénéfice
d'un corpus de taille raisonnable {--} environ cent
millions de mots pour un corpus général {--}, car certains
contextes peuvent apparaître dans ce type de corpus et être
considérés comme typiques d'un mot dans un sens
donné, sans qu'une relation syntaxique
n'unisse au premier chef le terme considéré et son
contexte. Une relation syntaxique, que nous appelons secondaire, peut
de ce fait être considérée comme pertinente, à condition que le
lien primaire ne soit interrompu qu'une seule fois.
Ainsi, on pourra considérer que dans les expressions
\textit{décrire un cercle} et \textit{décrire un arc de cercle}, le
terme \textit{cercle} peut également être considéré comme un
contexte de \textit{décrire}. Dans la seconde expression, le lien
secondaire entre \textit{décrire} et \textit{cercle}
n'est en effet interrompu qu'une
seule fois, puisque \textit{arc} est relié à la fois à
\textit{décrire} et à \textit{cercle} par une relation primaire.

 Les autres problèmes recouvrent la difficulté que peut éprouver
un système automatique pour appréhender la dimension flexionnelle
d'un grand nombre d'unités
lexicales. En effet, la méthode des contextes proches ne distingue
que des séquences de lettres, des graphies, et pas réellement les
unités lexicales. De ce fait, les expressions \textit{il a fait des
courses} et \textit{il fera des courses} seront traitées
séparément, alors que le sens lexical est identique. De même,
chaque forme d'un même nom ou adjectif sera
traitée séparément de toutes les autres, ce qui amène à
sous{}-estimer la typicité d'un contexte, ou même
à distinguer deux significations différentes pour
l'apparition d'un même contexte
dans le voisinage d'une même unité lexicale. Il
arrive bien sûr que la flexion soit choisie à dessein pour susciter
un changement de sens. Ainsi, une expression comme \textit{faire le
trottoir} n'est pas a priori porteuse
d'une dimension sémantique en rapport avec des
travaux publics routiers, contrairement à son pendant \textit{faire
les trottoirs}. Ces changements de sens sont cependant relativement
rares, et semblent plus souvent induits par des variations en
morphologie nominale qu'en morphologie verbale. Dès
lors, une analyse morphologique complète s'impose
comme la solution logique à ces difficultés. Un système
d'analyse morphologique automatique permet de
distinguer les catégories grammaticales des mots et
d'obtenir leur lemme tout en conservant la forme de
mot telle qu'elle apparaît dans le texte. De la
sorte, les différentes formes d'une même unité
lexicale peuvent être unifiées, tandis que sont distinguées des
unités différentes qui présentent une même graphie. Nous
proposons d'utiliser comme sommets des graphes des
unités lexicales sous forme de lemmes, et plus des formes de mots.
Cette préférence pour le lemme comme entité de référence
reste cependant paramétrable, essentiellement pour les substantifs,
conformément à la remarque faite plus haut. Par ailleurs,
l'analyse morphologique est également en mesure
d'éliminer automatiquement les mots{}-outils du
calcul des contextes typiques sans recourir à une suppression
statistique. En effet, leur sémantique, particulièrement faible,
n'a que très peu d'influence sur le
sens en contexte et n'est donc pas pertinente dans ce
type d'application. Du fait de cette élimination
plus ciblée, des unités lexicales particulièrement fréquentes
ne seront plus supprimées par erreur, tandis que des interjections,
prépositions et autres déterminants peu fréquents ne pollueront
plus les cartes sémantiques, comme c'était le cas
lorsque des techniques statistiques étaient utilisées.

 Ces différentes propositions nous ont amenés à revoir
entièrement les éléments d'information
manipulés par le modèle de représentation du sens dans
l'Atlas sémantique contextuel. En effet, ce sont non
seulement les relations entre sommets des cliques qui seront
modifiées, puisque de relations de proximité, elles deviendront
relations syntaxiques, mais aussi les sommets eux{}-mêmes, qui ne
seront plus des séquences de caractères ou des formes de mots, mais
des unités lexicales généralement sous forme de lemmes. Dès
lors, les éléments sur lesquels le modèle peut
s'appuyer pour construire les cliques correspondent
mieux par leur qualité aux caractéristiques des ressources
employées à l'origine pour le valider. La
construction des cliques peut de ce fait se conformer plus
complètement au modèle, et être calquée sur la méthode
utilisée dans le cadre de la construction de l'Atlas
sémantique synonymique. En effet, la transformation des données
manipulées par le modèle par rapport à la méthode des contextes
proches a rendu obsolètes les contraintes statistiques de limitation
du bruit. Ce sont donc des relations syntaxiques primaires ou
secondaires entre des lemmes ou formes de mots identifiées qui seront
réunies dans les cliques. Comme pour les autres applications, elles
seront ensuite disposées dans un espace géométrique
multidimensionnel, dont la projection sur un plan permettra de
visualiser les tendances de sens pour chaque unité lexicale
attestée dans le corpus exploité.

 Comme dans la méthode utilisée pour l'Atlas
contextuel, le dictionnaire ainsi produit sera représentatif du
corpus utilisé pour sa construction. L'ensemble du
lexique présent dans le corpus sera représenté dans la ressource
sémantique, et tous les sens attestés dans ce corpus trouveront
également leur reflet dans les cartes sémantiques. Nous insistons
à nouveau sur la nécessité de richesse et
d'excellence du corpus choisi, car la qualité du
dictionnaire sémantique sera fonction des facteurs de diversité et
d'étendue du corpus exploité. Il faut également
noter que plus le corpus utilisé est représentatif de la langue
étudiée, plus la ressource produite y sera également adaptée.
Le dictionnaire aura des caractéristiques plus générales si des
textes plus généraux ou moins ciblés sur un domaine apparaissent
dans le corpus.

\subsection{Représentation du sens et information en corpus}
\label{sec:infoCorpus}

 Construire une ressource lexico{}-sémantique grâce au modèle des
Atlas sémantiques à travers un corpus n'est pas
sans conséquences sur les possibilités de la ressource. En effet,
le dictionnaire contextuel que nous proposons de construire constituera
d'abord l'outil le plus adapté à
la description et à l'étude lexicale et
lexico{}-sémantique dudit corpus. Notamment,
l'ensemble du lexique présent dans les textes
apparaîtra dans la ressource, et tous les sens qui sont attestés
dans le corpus y seront également représentés, tandis que le
vocabulaire absent du corpus, ou les sens qui n'y sont
pas attestés, ne pourront figurer dans le dictionnaire. La figure~\ref{fig:regle}
illustre bien ce principe\footnote{La ressource que nous proposons de réaliser
n'est pas encore construite. Toutefois, certaines de
ses caractéristiques peuvent d'ores et déjà
être extrapolées sur la base de l'Atlas
contextuel.}. En effet, cette carte sémantique issue
de la ressource contextuelle, et donc constituée sans analyse
syntaxique ni morphologique, montre les différentes tendances de sens
de \textit{règle}. On remarque que des sens aussi courants que celui
de l'instrument de mesure ou celui des menstruations
n'apparaissent pas dans la carte sémantique. Il est
intéressant de noter que ces sens absent du dictionnaire ne figurent
pas non plus dans le corpus journalistique utilisé\footnote{Les cinq années du journal \textit{Le Monde} utilisées pour
constituer cette ressource sont pourtant un corpus très large de plus
de cent millions de mots. Il s'agit toutefois
d'un genre littéraire particulier, finalement assez
peu représentatif d'un état de langue naturel.}.

 D'autre part, la méthode que nous proposons permet
aisément de conserver un lien entre la représentation de chaque
tendance de sens des mots et les instances qui, dans le corpus, ont
permis de construire cette représentation, et en constituent dès
lors des exemples représentatifs. De plus, grâce à
l'analyse morphosyntaxique qui a été effectuée
sur le corpus, l'accès aux énoncés permet de
bénéficier également des schémas syntaxiques associés à la
tendance de sens sélectionnée, et d'en déduire
éventuellement une construction typique, un usage particulier ou un
schéma de sous{}-catégorisation plus ou moins précis. On pourrait
ainsi atteindre grâce à la carte sémantique de \textit{peindre}
un ensemble d'exemples où le verbe est
systématiquement rattaché à un groupe prépositionnel de
préposition \textit{en} dont la tête serait un nom de couleur,
dénotant ainsi le sens de \textit{couvrir de peinture de couleur}. On
se rapproche ainsi des spécificités et des qualités décrites
par \cite{MelcukAl95} pour la
constitution d'un \textit{Dictionnaire explicatif et
combinatoire}. En effet, le lien direct entre le sens
d'un mot et ses instances, à la fois syntaxiques et
lexicales, pourrait constituer un excellent outil
d'aide au traitement des vocables pour la construction
de dictionnaires tels que le \textit{DECFC}.

 Ce dictionnaire lexico{}-sémantique constitue donc la description ad
hoc et intégrale du vocabulaire présent dans le corpus et des sens
de ce vocabulaire qui y sont attestés, et les relations entre la
représentation sémantique et les énoncés qui ont servi à la
réaliser sont conservées et peuvent être suivies dans les deux
directions. De ce fait, cette ressource est un excellent outil de
navigation à travers l'information contenue dans le
corpus de référence. En effet, non seulement la carte sémantique
d'une unité lexicale fournit un accès immédiat
à l'ensemble des énoncés qui contiennent cette
unité dans un sens sélectionné, mais elle permet aussi
d'avoir accès à tous les exemples attestés
d'une expression syntaxiquement cohérente, puisque
la tendance de sens est exprimée par les contextes sémantiques
typiques de l'unité lexicale pour un sens
déterminé. Par ailleurs, il est possible
d'effectuer une navigation thématique à travers le
corpus, en passant de bribe en bribe, chacune d'entre
elles contenant une occurrence d'une même tendance
de sens, et reliées entre elles dans une carte sémantique. Il est
enfin envisageable d'exploiter un ensemble de cartes
sémantiques comme autant de clefs de recherche, et de sélectionner
finalement non pas les documents qui contiennent ces mots{}-clefs, mais
de restreindre les réponses aux documents qui contiennent les sens
désirés, réduisant ainsi le bruit parmi les réponses obtenues.
Au{}-delà de caractéristiques proposées également par les
dictionnaires classiques, la ressource que nous proposons comporte donc
un certain nombre de qualités intrinsèques qui en font un outil de
premier plan dans la perspective du traitement automatique de
l'information textuelle.

\subsection{Présentation pratique de l'approche}
\label{sec:pratique}

 Dans le cadre de la recherche présentée dans cet article, nous
avons réalisé un prototype qui permet de construire une ressource
lexico{}-sémantique conforme à notre approche à partir
d'un corpus textuel. Dans sa version actuelle, ce
prototype ne permet de gérer qu'une quantité
limitée de données, et ne dispose pas encore d'une
sortie graphique pour l'affichage des tendances de
sens dénotées par les cliques, comme c'est le cas
pour les Atlas synonymique et contextuel. Par ailleurs, deux ressources
sont nécessaires pour traiter le multilinguisme et le passage entre
deux langues. Deux corpus ont donc dû être constitués.

 Les corpus utilisés sont des ensembles de textes issus de la
sauvegarde de l'Encyclopédie \textit{Wikipédia}
datée de novembre 2006 dans ses instances française et anglaise.
Le texte en a été prétraité de manière à être
débarrassé de tout formatage typographique ainsi que des images,
tableaux et autres entités non textuelles qui risquaient de perturber
l'analyse. Les corpus sont constitués de 846
articles d'une taille minimale de 1500 mots, et dont
le titre est le même dans chacune des versions de
l'encyclopédie, de manière à disposer de
thématiques semblables malgré la taille réduite de
l'échantillon. Ils présentent toutefois des
différences, notamment dans le nombre total de mots
qu'ils contiennent, soit 2~329~811 mots français
et 3~303~498 mots pour l'anglais. Chaque corpus a
été entièrement soumis à une analyse morpho{}-syntaxique
automatique, respectivement par l'analyseur
\textit{SYNTEX} pour le Français \cite{BourigaultFabre00} et le
\textit{Stanford Parser} pour l'anglais \cite{KleinManning03}. 
Chacun de ces analyseurs présente la
caractéristique d'effectuer une analyse
morphologique et une désambiguïsation catégorielle préalable
à l'analyse syntaxique. Ils construisent ensuite un
ensemble de dépendances correspondant aux relations syntaxiques entre
les têtes des syntagmes repérés.

À l'issue de cette analyse, nous avons rejeté les
mots{}-outils, peu informatifs sur la typicité sémantique du
contexte, ainsi que toutes les dépendances qui impliquaient ces
mots{}-outils. Deux ensembles de dépendances ont ainsi été
formés, de 64~539 relations pour le français et de 84~769
relations pour l'anglais. Le lexique couvert \ est de
43~335 lemmes différents pour le français et de 55~348 pour
l'anglais. Ces ensembles ont été collectés dans
une base de données, et sont associées à leur énoncé
d'apparition de manière à établir un lien entre
un contexte et son usage authentique.

L'étape suivante consiste à constituer les
cliques, c'est{}-à{}-dire les graphes complets
formés par des ensembles de mots lorsqu'ils sont
reliés à tous les autres membres de la clique par une dépendance
syntaxique dans le corpus considéré. Les cliques considérées
sont de deux ordres : les cliques primaires, issues de relations
syntaxiques directes entre deux unités lexicales, et les cliques
secondaires, qui sont formées en considérant comme reliées entre
elles des unités lexicales ne comportant aucune dépendance en
commun, mais qui sont en relation avec une même unité lexicale
intermédiaire. L'algorithme de construction
dynamique des cliques en fonction de la lemmatisation ou non de
certaines catégories grammaticales est toujours en développement.
Toutefois, nous avons construit manuellement les cliques primaires
liées à une dizaine d'unités lexicales. Les
premiers tests menés sur ces cliques sont prometteurs, car ils font
apparaître notamment plus de contextes verbaux, qui étaient
précédemment minimisés voire gommés du fait de leur richesse
morphologique par l'approche de
l'Atlas contextuel.

L'étape ultérieure consistera à réaliser
l'interface graphique qui permettra de projeter ces
ensembles de cliques dans un espace sémantique de manière à
disposer d'une représentation sémantique des sens,
tout en distinguant les représentations issues de relations primaires
et secondaires.

\section{Corpus bilingues et information multilingue}
\label{sec:multilingue}

 En outre, le modèle des Atlas sémantiques se prête bien à une
utilisation avancée en contexte multilingue. En effet, S. Ploux a
montré combien il était aisé de mettre au moins partiellement en
correspondance la carte sémantique lexicale avec celle de sa
traduction, et ainsi d'indiquer dans quel sens une
unité lexicale peut être la traduction valide
d'une autre, et dans quel sens ce
n'est pas le cas \cite{PlouxJi03,Ploux07}.

 La méthode mise en {\oe}uvre vise à
l'enrichissement respectif des ressources synonymiques
de chacune des deux langues visées. Elle s'appuie
sur un simple dictionnaire de traduction qui est chargé de traduire
dans l'autre langue, et de toutes les manières
qu'il connaît, chaque unité lexicale présente
dans la carte sémantique de la langue source, de manière
d'une part à mettre en correspondance, lorsque
c'est possible, les différentes tendances de sens
des deux langues concernées, et d'autre part à
développer au maximum le vocabulaire dénotant les sens dans la
langue cible, de manière à enrichir la ressource synonymique de
cette langue.

 Notre proposition, dans le cadre de la construction
d'une ressource lexico{}-sémantique à finalité
de gestion de l'information textuelle, vise moins à
l'enrichissement des cartes sémantiques
qu'à leur mise en correspondance à travers les
langues traitées. Toutefois, outre le fait que la méthode de S.
Ploux a été conçue spécifiquement pour le modèle des Atlas
sémantiques, elle nous semble également capable de répondre à
nos besoins. Il s'agit cependant d'en
noter les exigences, en particulier en créant deux ressources
comparables, une pour chaque langue traitée. Pour ce faire, nous
devons disposer pour chaque langue d'un corpus de
taille raisonnable, c'est{}-à{}-dire susceptible de
contenir la majeure partie du lexique général courant, qui soit
représentatif non seulement du vocabulaire utilisé, mais
également de l'étendue sémantique de ce
vocabulaire. Par ailleurs, il faut idéalement que les thématiques
abordées par l'un le soient également par
l'autre avec un niveau de spécialisation globalement
comparable. Comme des corpus parallèles généraux de cette taille
ne sont pas disponibles et qu'il
n'est pas envisageable d'en créer
facilement et rapidement, l'exploitation de deux
instances de l'encyclopédie \textit{Wikipédia} de
langues différentes comme corpus comparables nous semble un choix
adapté à ces critères.

 En effet, ces instances de l'encyclopédie
constituent des corpus de grande taille dont la qualité
orthographique dépasse la moyenne des corpus collectés sur Internet
ou dans des journaux, du fait de l'intervention
immédiate de la part de nombreux lecteurs qui
s'instituent correcteurs voire contributeurs. Les
textes qu'elles contiennent se rapprochent également
des articles d'une encyclopédie générale,
abordant de ce fait de nombreux sujets plus ou moins spécialisés,
et qui donc sont susceptibles de contenir une part extrêmement large
du lexique de la langue pratiquée, ainsi que la plupart des
acceptions attestées de ce lexique. Ce point est capital pour la
généralité de la ressource produite, puisque
c'est la diversité des sens attestés dans le
corpus, et donc des contextes utilisés, qui permet une
représentation effective de ces sens dans une carte sémantique.
Enfin, chaque instance de \textit{Wikipédia} comporte des articles
qui sont le pendant d'articles présents dans une ou
plusieurs autres instances : on trouvera par exemple un article
traitant de \textit{Louis~XIV} à la fois dans la \textit{Wikipédia}
française et anglaise, sans que l'article
d'une langue soit la traduction de
l'article dans l'autre langue. Cette
communauté de sujets abordés, pour partie du moins, est également
un atout non négligeable, car il fournit la certitude que certaines
thématiques, et donc certaines parties du lexique, seront communes
à chaque langue considérée, et donc que la nature des langues
traitées aura une certaine cohérence. Cela ne pourrait être le
cas en utilisant deux corpus trop distincts, par exemple un corpus
médical dans une langue et un corpus agronomique dans
l'autre.

 Nous proposons donc de réaliser, pour le français et pour
l'anglais (mais d'autres langues sont
possibles), des ressources lexico{}-sémantiques sur base de leur
\textit{Wikipédia} respective, en effectuant une analyse
morphosyntaxique de chaque article et en construisant pour chaque
unité lexicale identifiée l'ensemble des cliques
disponibles à projeter sous forme de carte sémantique. Une
traduction, même partielle, des contextes syntaxiques permettra de
faire correspondre autant que possible les cartes sémantiques. Les
lacunes inévitables devraient souvent être comblées, dans la
mesure où d'autres contextes typiques dénotant la
même tendance de sens ont pu trouver leur contrepartie dans
l'autre langue. Ainsi, la carte sémantique de
\textit{poisson} dans le sens {\guillemotleft} poisson
d'agrément {\guillemotright} devrait présenter des
contextes typiques comme \textit{aquarium} ou \textit{bassin}. Pour
l'anglais, on en trouvera généralement la
traduction : \textit{aquarium}, \textit{pond}. Mais le correspondant
anglais \textit{gold} du contexte typique \textit{rouge}
n'en est pas la traduction. Pourtant, les instances de
\textit{poisson rouge} seront bel et bien reliées dans la carte
sémantique à la tendance de sens correspondant au poisson
d'agrément, et il en va de même pour le
\textit{gold fish} anglais. Or ces tendances de sens se répondent
lorsque les cartes sémantiques française et anglaise sont mises
en correspondance. Les documents traitant de \textit{poisson rouge}
dans une langue seront donc reliées à ceux qui parlent de
\textit{gold fish} dans l'autre, et inversement. Le
passage d'une information dans une langue à son
expression dans l'autre est désormais possible, sans
pour autant maîtriser toutes les étapes de la traduction.

 Le prototype que nous avons présenté à la section 3.3., malgré
le caractère limité des données qu'il est
actuellement capable de traiter, nous a permis de tester la mise en
{\oe}uvre de la méthode que nous proposons. Les ressources
construites dans chaque langue sont constituées, pour chaque unité
lexicale considérée, d'un ensemble de cliques qui
permettront de représenter ses tendances de sens sous forme
d'une carte sémantique.
L'opération de traduction est effectuée grâce
à une ressource interne du LIMSI, un dictionnaire bilingue
français{}-anglais construit progressivement dans le cadre
d'autres projets, comportant environ 40~000 entrées
et 250~000 traductions pour chaque langue. Aucun effort
n'est fait à ce stade pour trouver la meilleure
traduction de l'expression de départ dans la langue
correspondante en fonction de son sens original : la sélection de la
carte d'une entrée dans la langue{}-source provoque
dans un premier temps la sélection des cartes de toutes les
traductions de cette entrée dans la langue{}-cible proposées par le
dictionnaire.

 Que ce soit dans la langue{}-source ou dans la langue{}-cible, ce sont
donc des ensembles de cliques comportant des contextes syntaxiques des
entrées correspondantes qui sont sélectionnés. Le dictionnaire
permet ainsi de rechercher la ou les traductions de chaque contexte
présent dans une clique de la carte dans la langue{}-source, et de
rechercher leur présence dans des cliques présentes dans une ou
plusieurs des cartes de la langue{}-cible. Certaines cliques de la
langue{}-source sont ainsi massivement traduites dans la
langue{}-cible, tandis que d'autres le sont beaucoup
moins, et ne peuvent donc être considérées comme la
correspondance d'une clique
d'origine. Il arrive également que les composants
d'une clique trouvent leurs traduction, mais
dispersées dans différentes cliques qui se recouvrent peu. Ces
traductions ne sont pas non plus à prendre en compte.

 Nous avons donc effectué des tests manuels à partir du prototype
déjà construit, et notamment sur le terme français
\textit{poisson} dont la traduction est \textit{fish. }Sur notre
échantillon, le contexte rouge apparaît dans deux cliques
françaises, et le mot \textit{gold} dans une clique anglaise. Or on
retrouve dans les deux cliques françaises et dans la clique
anglaise quatre contextes qui se répondent en traduction :
\textit{petit}, \textit{tropical}, \textit{aquarium} et
\textit{nourrir} pour le français, \textit{small},
\textit{tropical}, \textit{aquarium} et \textit{feed} pour
l'anglais. Les cliques françaises sont
constituées de sept et neuf contextes, et la clique anglaise de sept.
Depuis les phrases parlant de \textit{poisson rouge} dans le corpus
français, la sélection des cliques françaises qui contiennent
le mot \textit{rouge} permet donc d'avoir accès à la clique
anglaise qui contient \textit{gold} et donc aux énoncés traitant de
\textit{gold fish} sans que la traduction ait été nécessaire.
Bien entendu, ce test ne peut pas être considéré comme
représentatif, ni de la langue, ni d'un corpus, et
demande à être systématisé sur un plus grand volume de
données. Il est toutefois encourageant et va dans le sens de nos
hypothèses.

\section{Conclusion}
\label{sec:Conclusion}

 Grâce à son application aux relations de synonymie issue de
dictionnaires et de proximité contextuelle tirée de corpus de
grande taille, le modèle des Atlas sémantiques a fait montre de ses
qualités pour la description de la sémantique lexicale. En
particulier, son exploitation contextuelle a indiqué sa capacité
à dénoter le sens attesté dans un corpus sans prendre en compte
un lien d'ordre sémantique, et à en représenter
l'information. D'autre part, les
cartes sémantiques de représentation du sens fournies par ce
modèle peuvent aisément être mises en correspondance entre
langues en utilisant un simple dictionnaire de traduction. Cependant,
si le modèle est bien validé par différentes expériences et
évaluations, sa mise en {\oe}uvre souffre de plusieurs imperfections,
essentiellement liées à une méthode qui ne prend pas en compte la
caractéristique langagière des textes.

 Nous proposons de réaliser une ressource lexico{}-sémantique
basée sur le modèle des Atlas sémantiques et
l'utilisation de corpus, mais en mettant en {\oe}uvre
une analyse morphosyntaxique pour établir les cliques rassemblant les
unités lexicales. Cette ressource lexicale constituera de ce fait un
descriptif \textit{ad hoc} du contenu du corpus et, à ce titre, un
instrument idéal pour en appréhender et traiter
l'information et pour naviguer sémantiquement à
travers ses contenus. Par ailleurs, la préservation du lien entre les
cartographies sémantiques et les énoncés qui ont servi à les
réaliser donne la possibilité d'accéder aux
exemples authentiques d'utilisation
d'une unité lexicale donnée dans son sens
désiré contenus dans le corpus. L'analyse
morphosyntaxique préalable du corpus permet également de déduire
sa construction syntaxique d'usage, voire un schéma
de sous{}-catégorisation.

 En outre, nous suggérons d'exploiter parallèlement
des corpus similaires de grande taille et de langues distinctes, à
savoir différentes instances linguistiques de
l'encyclopédie libre \textit{Wikipédia}, pour
créer des ressources comparables dans des langues différentes, dont
les représentations sémantiques peuvent être mises en
correspondance. La navigation thématique, possible dans chaque corpus
individuellement, devrait dépasser la barrière des langues grâce
aux correspondances de cartographies sémantiques de chaque corpus.
Ces représentations permettent en effet d'associer
des énoncés portant sur un même sujet sans distinction de langue,
car le dictionnaire bilingue assure la traduction pour
l'unité lexicale recherchée et pour la plupart de
ses contextes dans les cartes sémantiques, ce qui permet de
restreindre les textes fournis à l'apparition
d'un sens sélectionné. Et quand le dictionnaire
bilingue échoue à fournir une traduction à certaines contextes
typiques d'une unité lexicale, la navigation
thématique interlangue reste possible, car les contextes traduits
appartenant à la même tendance de sens permettent
d'identifier dans une autre langue la tendance de sens
correspondante, et donc les bribes de textes qui lui sont liées. Les
tests effectués grâce à un premier prototype sur un ensemble
textuel moyen sont encourageants et vont dans le sens de notre
hypothèse.

 Mais au{}-delà de la perspective de réaliser un outil lexical
objectif et exhaustivement représentatif du contenu
d'un corpus, voire de fournir un instrument de
navigation textuelle et thématique interlangue à travers les
informations présentes dans le corpus, cette ressource ouvre
plusieurs pistes tant dans l'étude de la langue
qu'en applications pratiques. En effet, nous projetons
de nous pencher sur la nature même de l'unité
lexicale et sur son rapport au sens en utilisant pour ce faire les
contextes typiques et leur distribution dans les cartes sémantiques.
Dans cette perspective, c'est essentiellement
l'étude du problème des expressions à mots
multiples et de leur éventuelle lexicalisation qui constituera notre
base de réflexion. Nous pensons également pouvoir établir
automatiquement des liens de rapport sémantique entre unités
lexicales, classiques ou inédits, grâce aux similitudes et
divergences de leurs cartes sémantiques respectives.

 D'un point de vue plus applicatif, la ressource
proposée peut évidemment être utilisée pour aider à
l'élaboration de dictionnaires plus proches de la
forme traditionnelle de ces ouvrages, ou nécessitant certaines des
informations que notre approche est apte à fournir : un lexique et
une diversité sémantique représentatifs du corpus
sélectionné, un accès direct à une collection
d'exemples authentiques d'usages des
unités lexicales dans chaque sens attesté, une association de
chaque sens de toutes les unités avec sa catégorie grammaticale,
ainsi qu'avec un ou plusieurs schémas syntaxiques ou
de sous{}-catégorisation, etc. La relation établie entre les
différents sens des unités lexicales et des énoncés non
seulement dans la même langue, mais également avec des bribes de
texte dans d'autres langues permet également
d'envisager l'utilisation des
ressources de langues différentes pour aider à la rédaction de
dictionnaires bilingues, ou même pour constituer une forme de
mémoire de traduction, où des expressions regroupant une unité
lexicale et certains de ses contextes trouveront leur correspondance
dans d'autres langues grâce à
l'utilisation des cartes sémantiques et des corpus.

 Comme on peut le voir, les qualités du modèle des Atlas
sémantique ainsi que les apports d'une approche
mixte par corpus permettent d'envisager non seulement
une étude approfondie de la langue, mais aussi des instruments
performants en gestion de l'information textuelle,
mais encore des outils nécessaires dans des domaines aussi variés
que la création de dictionnaires et la réalisation de mémoires de
traduction.

\section*{Remerciements}
\label{sec:Remerciements}

 Nous tenons à exprimer notre plus vive gratitude à Hyungsuk Ji,
concepteur du modèle ACOM (Atlas sémantique contextuel), grâce
à qui cette recherche a pu voir le jour. Une partie de la réflexion
présentée ci{}-dessus a été menée lors d'un
projet financé par le programme TCAN (Traitement des Connaissances,
Apprentissage et NTIC) du CNRS.

\bibliographystyle{apalike-fr}

\bibliography{\mabiblio}

\end{document}